\documentclass[letterpaper]{article} 

\usepackage{aaai23}  
\usepackage{times}  
\usepackage{helvet}  
\usepackage{courier}  
\usepackage[hyphens]{url}  
\usepackage{graphicx} 
\usepackage{natbib}  
\usepackage{caption} 
\DeclareCaptionStyle{ruled}{labelfont=normalfont,labelsep=colon,strut=off} 
\usepackage{algorithm}
\usepackage{algorithmic}

\usepackage[dvipsnames]{xcolor}
\usepackage{booktabs}       
\usepackage{tabularx}
\usepackage{fancybox}
\usepackage{multirow}
\usepackage{soul,colortbl}
\usepackage{hhline}
\usepackage{arydshln}
\usepackage{fancyhdr,graphicx,amsmath,amssymb}
\definecolor{tblue}{RGB}{23, 73, 146}
\definecolor{bblue}{RGB}{147, 205, 221}
\definecolor{Gray}{gray}{0.9}
\definecolor{dred}{RGB}{192, 0, 0}
\newcommand{\RN}[1]{%
	\textup{\lowercase\expandafter{\it \romannumeral#1}}%
}
\usepackage{pifont}
\newcommand{\cmark}{\ding{51}}%
\newcommand{\xmark}{\ding{55}}%
\usepackage{newfloat}
\usepackage{listings}
\floatstyle{ruled}
\newfloat{listing}{tb}{lst}{}
\floatname{listing}{Listing}

\title{Adversarial Learning Meets Cooperative Learning}
\title{CoopInit: Initializing Generative Adversarial Networks via Cooperative Learning}
\author{
  Yang Zhao,
  Jianwen Xie,
  Ping Li\\
}
\affiliations{
    Cognitive Computing Lab\\
    Baidu Research\\
    10900 NE 8th St. Bellevue, WA 98004, USA\\
   \ \{yangzhao.eric, jianwen.kenny, pingli98\}@gmail.com
}

\begin{document}

\maketitle

\begin{abstract}
Numerous research efforts have been made to stabilize the training of the Generative Adversarial Networks (GANs), such as through regularization and architecture design. However, we identify the instability can also arise from the fragile balance at the early stage of adversarial learning. This paper proposes the CoopInit, a simple yet effective cooperative learning-based initialization strategy that can quickly learn a good starting point for GANs, with a very small computation overhead during training. The proposed algorithm consists of two learning stages: (i) Cooperative initialization stage: The discriminator of GAN is treated as an energy-based model (EBM) and is optimized via maximum likelihood estimation (MLE), with the help of the GAN's generator to provide synthetic data to approximate the learning gradients. The EBM also guides the MLE learning of the generator via MCMC teaching; (ii) Adversarial finalization stage: After a few iterations of initialization, the algorithm seamlessly transits to the regular mini-max adversarial training until convergence. The motivation is that the MLE-based initialization stage drives the model towards mode coverage, which is helpful in alleviating the issue of mode dropping during the adversarial learning stage. We demonstrate the effectiveness of the proposed approach on image generation and one-sided unpaired image-to-image translation tasks through extensive experiments. 
\end{abstract}

\section{Introduction}\label{sec:intro}
Generative modeling has proven to be an effective approach in many scenarios, e.g., image synthesis~\cite{xie2016theory,xie2018cooperativePAMI,brock2018large,karras2019style,zhao2021learning} and sequence generation~\cite{tulyakov2018mocogan,yu2017seqgan}. One of the most popular and powerful generative frameworks to date is the Generative Adversarial Network (GAN)~\cite{goodfellow2014generative}, which defines a mini-max game seeking a Nash equilibrium between a discriminator and a generator. Despite the recent successes of GANs in modeling complex high-dimensional distributions and generating realistic images~\cite{brock2018large,karras2020analyzing}, their training suffers from instability issues due to alternating parameter update~\cite{heusel2017gans}, the sensitivity to the hyper-parameter choices~\cite{salimans2016improved} and mode collapse issues~\cite{AroraRZ18}. To alleviate these issues, several techniques have been proposed, including gradient penalty~\cite{arjovsky2017wasserstein,mescheder2018training}, spectral normalization~\cite{miyato2018spectral}, discriminator bottleneck~\cite{zhao2020feature} and data augmentation~\cite{karras2020training}. In contrast, Generative Cooperative Networks (CoopNets)~\cite{xie2018cooperativePAMI} are another class of generative framework that jointly trains a descriptor and a generator, which has been successfully applied to image synthesis~\cite{XieZL21,XieZLL22}, 3D generation~\cite{xie2020generative}, supervised conditional learning~\cite{xie2019cooperative}, salient object prediction~\cite{ZhangXZB22}, unpaired image-to-image translation~\cite{xie2021CycleCoop}, and image hashing~\cite{XieHash22}.
Unlike GANs, CoopNets are optimized through cooperative maximum likelihood estimation (MLE). The descriptor, essentially a generative energy-based model (EBM)~\cite{xie2016theory,NijkampHZW19,du2019implicit}, incorporates the Stochastic Gradient Markov Chain Monte Carlo (SG-MCMC) to approximate the data distribution. The generator is an amortized sampler that simultaneously chases the descriptor towards the data distribution. The MLE-based learning scheme is often more stable and does not suffer from mode collapse issues. However, the training of CoopNets relies on an expensive MCMC sampler. It has also been suggested ~\citep{xie2020representation} that likelihood-based generative models tend to generate blurry images because they are obliged to fit all the major modes of the empirical data distribution. If they cannot fit the modes closely, they interpolate the major modes.

\begin{figure*}[!t]
    \centering
    \includegraphics[width=0.98\textwidth]{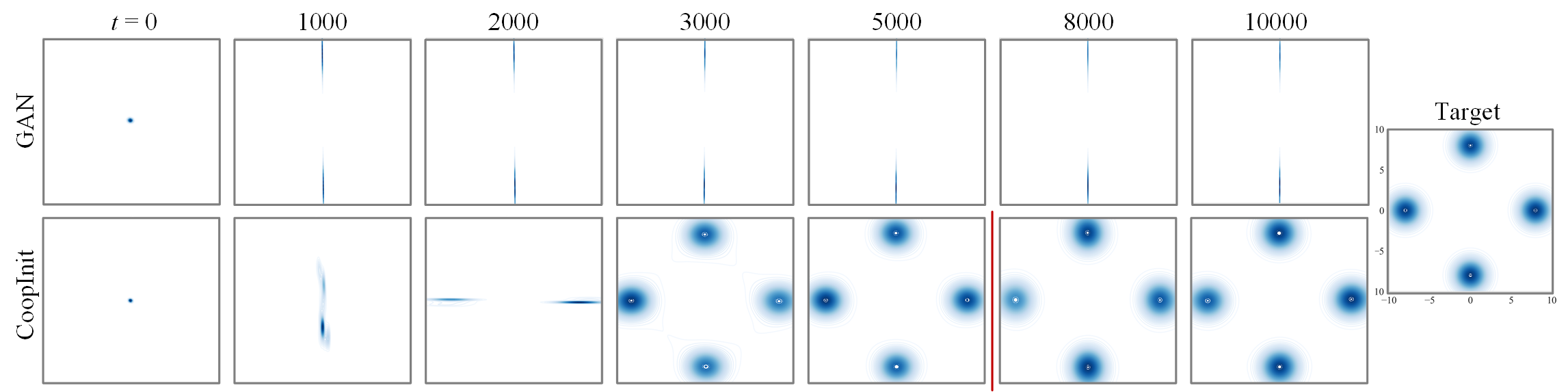}
    \caption{Comparison between standard GAN training (top row) and the GAN training with proposed CoopInit strategy (bottom row) trained on a 2D synthetic data distribution shown in the rightmost plot. Each column displays the generated distributions by the two methods at a different time step during training. The red vertical line displayed in the bottom row separates the cooperative initialization stage (left part) and the adversarial finalization stage (right part). The standard GAN training fails to converge to the target distribution because it encounters a severe mode collapse issue. In contrast, the GAN training using the CoopInit can benefit from the initial cooperative learning which helps overcome the mode collapse issue. }
    \label{fig:intro-cmp}
\end{figure*}

In this work, we aim to combine adversarial learning and cooperative learning to create stable, efficient, and powerful generative models. We propose a novel approach that leverages the strengths of both learning schemes. We first demonstrate that CoopNets and GAN can share network structures, so that we can treat them as one framework conveniently. Specifically, the discriminator in GAN can be transformed into the descriptor in CoopNets, and vice versa.
In other words, a bottom-up ConvNet, which plays the role of energy function of a descriptor in cooperative learning, can take a new role of discriminator in adversarial learning. 

Moreover, we propose a hybrid and effective strategy to train generative models. 
Specifically, the proposed framework consists of two networks, a bottom-up network $D_{\theta}$ parameterized by $\theta$ and a top-down network $G_{\phi}$ parameterized by $\phi$. Our hybrid learning algorithm includes two stages, each of which corresponds to a different learning scheme: 
At the first stage (\emph{cooperative initialization}): we train $D_{\theta}$ and $G_{\phi}$ in the cooperative learning scheme, where $D_{\theta}$ serves as an expressive EBM, to encourage mode coverage; at the second stage (\emph{adversarial finalization}): we continue to train $D_{\theta}$ and $G_{\phi}$ in the adversarial learning scheme, with parameters $\{\theta, \phi\}$ initialized from the first stage. The cooperative initialization stage only takes a small amount of time at the very beginning of the whole learning process. Intuitively, we first allow the stable cooperative learning to capture the majority of the mode structure of the data distribution to avoid mode collapse or dropping, and then the subsequent adversarial learning focus on refining the synthesis details through mode chasing. We call the proposed method the CoopInit, which can be considered a learning-based initialization approach for GAN training. We demonstrate the effectiveness of CoopInit through a synthetic experiment in Figure~\ref{fig:intro-cmp}. We highlight our main contributions below:
\begin{itemize}
\item We are the first to study how to combine the adversarial learning (i.e., GAN) and the cooperative learning (i.e., CoopNets) for generative modeling. It stabilizes and improves the adversarial training by firstly performing likelihood-based cooperative learning for initialization.
\item  We conduct extensive experiments for model analysis and ablation study in order to understand the behavior of the proposed learning algorithm. 
\item We demonstrate that the proposed training strategy can outperform previous CoopNets and GANs, and obtain state-of-the-art performance in image generation benchmarks and one-sided image translation benchmarks.
\end{itemize}
The rest of the paper is organized as follows: In Section~\ref{sec:preliminary}, we present preliminaries of adversarial learning and cooperative learning. Section~\ref{sec:method} describes the proposed learning framework and its theoretical understanding in detail. In Section~\ref{sec:work}, we present prior arts that are related to our model. In Section~\ref{sec:experiments}, we validate the proposed method via extensive experiments. Finally, in Section~\ref{sec:conclusion}, we conclude our~work.

\section{Preliminaries}\label{sec:preliminary}

The generator, denoted by $G_{\phi}$, seeks to transform a prior distribution of latent space $z \sim p(z)$, via a top-down network, into a distribution that can approximate the ground truth data distribution $p_{\text{data}}(x)$. The generator $G_{\phi}$ can pair up with either a discriminator for adversarial training or a descriptor for cooperative training, both of which can be parameterized by a bottom-up network $D_{\theta}$. $\theta$ and $\phi$ are parameters.

\subsection{Adversarial Learning}\label{sec:gan}
GANs~\cite{goodfellow2014generative} define a minimax game between the discriminator $D_{\theta}$ and the generator $G_{\phi}$. The generator $G_{\phi}$ tries to generate realistic examples to fool the discriminator $D_{\theta}$ whereas the discriminator $D_{\theta}$ aims to distinguish between the generated examples $G_{\phi}(z)$ where $z \sim p(z)$ and the real data examples $x \sim p_{\text{data}}(x)$. ~\citet{goodfellow2014generative} proposed an adversarial loss, given by
\begin{align} 
    \mathcal{L}^{\text{adv}}= \mathbb{E}_{p_{\text{data}}(x)} [\log D_{\theta}(x)] - \mathbb{E}_{ p(z)} [\log (1 - D_{\theta}(G_{\phi}(z)))]. \notag
\end{align} 
The generator tries to minimize  $\mathcal{L}^{\text{adv}}$ while the discriminator tries to maximize $\mathcal{L}^{\text{adv}}$. In practice, to circumvent the vanishing gradient issues caused by a saturated discriminator, the generator is instead trained to maximize $\mathbb{E}_{p(z)} [\log D_{\theta}(G_{\phi}(z))]$. This non-saturating (NS) loss is used in a series of StyleGAN models~\cite{karras2019style,karras2020analyzing,karras2020training} and related works~\cite{choi2020stargan,pidhorskyi2020adversarial}. Wasserstein distance~\cite{arjovsky2017wasserstein} (WAS) is also a standard divergence used to train GANs. However, the introduced Lipschitz constraint in WGAN usually relies on weight clipping and is sensitive to parameters. A follow-up WGAN-GP~\cite{gulrajani2017improved} instead proposes to add a gradient penalty (GP) to the WAS for enforcing the Lipschitz continuity. A notable example of WAS-GP is the ProgressiveGAN~\cite{karras2017progressive}.
The hinge loss~\cite{lim2017geometric,tran2017deep} (Hinge) is another common objective used to train GANs, for example in BigGAN~\cite{brock2018large} and SN-GAN~\cite{miyato2018spectral}. 
We will evaluate the three aforementioned variants of adversarial loss in the experiments.
\begin{figure*}[!htbp]
    \centering
    \includegraphics[width=0.8\textwidth]{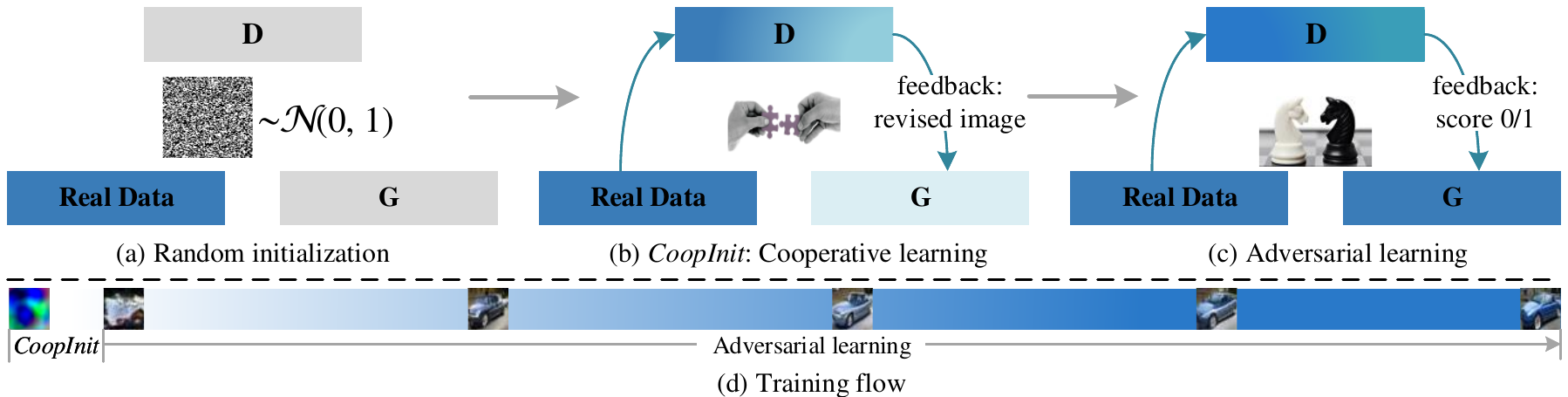}
    \caption{An illustration of the CoopInit technique for improving GAN training. \textbf{D}: discriminator or descriptor. \textbf{G}: generator.}
    \label{fig:model}
\end{figure*}

\subsection{Cooperative Learning}
In contrast to GANs, CoopNets apply a cooperative learning strategy to train the generator $G$ and the descriptor $D$ simultaneously via MCMC teaching. The descriptor $D$ is essentially an EBM \cite{xie2016theory}, which is defined as:
\begin{align}
\label{eq:unnorm-prob}
p_\theta(x) = \frac{1}{Z(\theta)}\exp [D_\theta(x)],
\end{align}
\par \vspace{-1mm}
\noindent where $D_\theta(x)$ is the negative energy function defined on data domain and $Z_\theta$ is the intractable normalizing constant. To learn the descriptor, we seek to maximize the log-likelihood:
\begin{align}\label{eq:ll}
    \mathcal{L} = \mathbb{E}_{ p_{\text{data}}(x)} [ \log p_\theta(x)],
\end{align}
\par \vspace{-1mm}
\noindent which is equivalent to minimizing the Kullback-Leibler divergence $\mathsf{KL}(p_{\text{data}}(x)||p_\theta)$.  Its derivative is given by
\begin{align}\label{eq:cd-update}
    \nabla_\theta \mathcal{L} &= \mathbb{E}_{p_{\text{data}}(x)} [\nabla_\theta D_\theta(x)] - \mathbb{E}_{p_{\theta}(x)} [\nabla_\theta D_\theta(x)] \notag\\
    &\approx \frac{1}{n} \sum_{i=1}^{n} \nabla_\theta D_\theta(x_i) -\frac{1}{n} \sum_{i=1}^{n} \nabla_\theta D_\theta(\tilde{x}_i),
\end{align}
where $\{x_i\} \sim p_{\text{data}}(x)$ are observed examples and $\{\tilde{x}_i \}\sim p_{\theta}(x)$ are synthesized examples generated via MCMC, such as Langevin dynamics~\cite{zhu1998grade} that iterates the following step 
\begin{align}\label{eq:LD}
x_{t+1}=x_{t} + \eta \nabla_x D_\theta(x_t) +\epsilon_t, \epsilon_t \sim \mathcal{N}(0, \sqrt{2\eta}I),
\end{align}
with $t$ indexing the Langevin time step and $x_{t=0}$ being initialized by random noise. $\eta$ is a~hyperparameter for Langevin step size. 
In high dimensional modeling cases, the MCMC can be expensive and difficult to converge. However, CoopNets can improve the sampling by using a generator $G_{\phi}$ to generate initial synthesized examples to initialize a finite-step MCMC that samples and trains the descriptor $D_{\theta}$.  The generator serves as an amortized sampler for the descriptor. The generator $G_{\phi}$ updates its parameters by directly learning from the synthesized examples produced by the MCMC, which is called MCMC teaching.   
The descriptor learns from the difference between the MCMC outputs and the training examples, while the generator learns from how the descriptor revises the initial outputs. Algorithm~\ref{alg:coopnet} presents one iteration of the cooperative learning. 

  \begin{center}
    \begin{minipage}{0.9\linewidth}
        \scriptsize
\begin{algorithm}[H]
  \caption{Cooperative Learning}
  \label{alg:coopnet}
\begin{algorithmic}
\REQUIRE{descriptor $D_\theta$, generator $G_\phi$, Langevin dynamics step size $\eta$, number of Langevin steps $T$.}
    \STATE { {\small \color{RoyalBlue} $\#  \mathtt{~Step~G1:~Generate~initial~examples~\hat{x}}$}} 
    \STATE {$\hat{z}_i \sim p(z), \hat{x}_i = G_\phi(\hat{z}_i)$}
    \STATE {{\small \color{RoyalBlue} $\#  \mathtt{~Step~D1:~Revise~\tilde{x}~for~{\it T}~steps~via~LD}$}}
    \STATE { Initialize $\tilde{x}_i=\hat{x}_i$}
    \FOR{$1$ to $T$}
    \STATE {$\tilde{x}_i \leftarrow \tilde{x}_i + \eta \nabla_{x} D_\theta(\tilde{x}_i) + \epsilon, \epsilon \sim \mathcal{N}(0, \sqrt{2\eta}I)$}
    \ENDFOR
    \STATE {{\small \color{RoyalBlue} $\#  \mathtt{~Step~D2:~Train~descriptor~}$$D_\theta$}}
    \STATE{ Train~$D_\theta$ with gradient descent via Eq.~\eqref{eq:cd-update}}
    \STATE { {\small \color{RoyalBlue} $\#  \mathtt{~Step~G2:~Train~generator~}$$G_\phi$}}
    \STATE{ Train $G_\phi$ with gradient descent on $\frac{1}{n}\sum_{i=1}^{n}||\tilde{x}_i - G_{\phi}(\hat{z}_i)||^2$}
\end{algorithmic}
\end{algorithm}
\end{minipage}
\end{center}
  \begin{center}
    \begin{minipage}{0.9\linewidth}
        \scriptsize
\begin{algorithm}[H]
  \caption{Training a GAN with CoopInit}
  \label{alg:coopinit}
\begin{algorithmic}
\REQUIRE{descriptor $D_\theta$ , generator $G_\phi$, number of examples consumed by cooperative initialization $N_{\text{coop}}$, and number of examples consumed by adversarial finalization $N_{\text{adv}}$, batch size $n$.} \\
\STATE {$N \leftarrow 0$}\\
{\small \color{RoyalBlue} $\#  \mathtt{~Train~D~and~G~as~CoopNets}$}
  \WHILE{$N \leq N_{\text{coop}}$}
        \STATE {Run Algorithm~\ref{alg:coopnet}} to update $D_\theta$ and $G_\phi$
        \STATE {$N \leftarrow N + n$}
  \ENDWHILE\\
{\small \color{RoyalBlue} $\#  \mathtt{~Train~D~and~G~as~GAN}$}
  \WHILE{$N \leq (N_{\text{adv}} + N_{\text{coop}})$}
        
        \STATE {Update $D_\theta$ and $G_\phi$ according to $\mathcal{L}^{\text{adv}}$} \\
        
        \STATE {$N \leftarrow N + n$}
  \ENDWHILE
\end{algorithmic}
\end{algorithm}
\end{minipage}
\end{center}

\section{CoopInit: A Strategy to Initialize GAN Training via Cooperative Learning}\label{sec:method}

\subsection{Proposed Framework}
Our generative learning framework, shown in Figure~\ref{fig:model}, integrates CoopNets and GAN, enabling us to smoothly switch between cooperative learning and adversarial learning. The proposed method begins with limited iterations of cooperative learning and then switches to adversarial learning until completion. We monitor the training progress using the number of training examples processed by the model. Specifically, 
We use $N_{\text{coop}}$ and $N_{\text{adv}}$ to represent the numbers of training examples consumed during cooperative learning and adversarial learning, respectively. The full description of training a GAN with CoopInit is shown in Algorithm~\ref{alg:coopinit}. In this paper, we always ensure that $N_{\text{coop}} / N_{\text{adv}} < 3$ to keep the computational overhead from MCMC negligible.

One might question why we don't simply use a combined objective of cooperative and adversarial learning. However, in practice, we have found that their compatibility is poor, resulting in an FID~\cite{heusel2017gans} of approximately 35 for image generation on CIFAR-10~\cite{Krizhevsky09learningmultiple} dataset  using both cooperative and adversarial learning simultaneously.
The cooperative learning leads to an MLE solution, which corresponds to a forward Kullback–Leibler (KL)-divergence, while the adversarial learning corresponds to Jensen–Shannon divergence, which involves a reverse KL-divergence. Thus, learning the models using these two objectives at the same time might lead to undesirable outcome due to incompatibility. Although both CoopNets and GAN use an alternating optimization procedure between $D_{\theta}$ and $G_{\phi}$, the key difference between cooperative and adversarial learning lies in that CoopNets uses MLE but GAN uses an adversarial loss. Further analysis of their optimization procedures reveals the following:

(i)
The role of $D_{\theta}$ differs in the optimization of GAN and CoopNets. In GAN, $D_{\theta}$ functions as a classifier that distinguishes between real data and generated data. In CoopNets, $D_{\theta}$ is a score (negative energy) function that assigns lower scores to generated data and higher scores to real data. 

(ii)
The objective of $G_{\phi}$ differs in the optimization of GAN and CoopNets. In GAN, $G_{\phi}$ is optimized by fooling $D_{\theta}$ into believing that generated examples are real. On the other hand, in CoopNets, $G_{\phi}$ is optimized by moving the generator's distribution towards the descriptor's distribution. 

\subsection{Theoretical Understanding}
We use $\mathcal{M}_{\theta}$ to denote the $T$-step MCMC transition kernel of the descriptor $p_{\theta}$. We also use $\mathcal{M}_{\theta}q_{\phi}$  to denote the marginal distribution obtained by
running the Markov transition $\mathcal{M}_{\theta}$ starting from the generator $q_{\phi}$. At each iteration $t$, the cooperative learning algorithm alternates the following two steps: 
(i) Update $\theta$: it  learns $\theta$ by minimizing
\begin{eqnarray}\label{eq:convergence1}
  \mathsf{KL}(p_{\rm data}{\parallel}p_{\theta}) - \mathsf{KL}(\mathcal{M}_{{\theta}^{(t)}} q_{{\phi}^{(t)}}{\parallel} p_{\theta})
\end{eqnarray} \par \vspace{-1mm}
\noindent over $\theta$, which is a modified contrastive divergence~\cite{xie2018cooperativePAMI} for the energy-based model $p_{\theta}$, and (ii) Update $\phi$: it learns $\phi$ by minimizing
\begin{eqnarray}\label{eq:convergence2}
  \mathsf{KL}(\mathcal{M}_{{\theta}^{(t)}} q_{{\phi}^{(t)}}{\parallel} q_{\phi})
\end{eqnarray}\par \vspace{-1mm}
\noindent over $\phi$. In an idealized situation where the generator $q_{\phi}$ has infinite capacity, the objective in Eq.~(\ref{eq:convergence2}) can be minimized to zero, which means that $q_{\phi}$ becomes the stationary distribution of $\mathcal{M}_{\theta}$, i.e., $\mathcal{M}_{\theta}q_{\phi}=q_{\phi}$, or equivalently $p_{\theta}=q_{\phi}$ (the generator has caught up with the descriptor and become an amortized sampler for the descriptor). Once this happens, the second
KL-divergence in Eq.~(\ref{eq:convergence1}) vanishes, because $\mathsf{KL}(\mathcal{M}_{{\theta}} q_{{\phi}}{\parallel} p_{\theta})=\mathsf{KL}(q_{\phi}{\parallel} p_{\theta})=0$. Then the learning of $\theta$ becomes maximum likelihood estimate that minimizes only the first KL-divergence $\mathsf{KL}(p_{\rm data}{\parallel}p_{\theta})$ in Eq.~(\ref{eq:convergence1}). Since $q_{\phi}$ chases $p_{\theta}$ toward $p_{\text{data}}$, the learning of $\phi$ is also a maximum likelihood estimate.  

In the second stage of the proposed algorithm, known as adversarial finalization, 
we continue to train $G_\phi$, which is initialized by the cooperative learning, to further refine its ability to capture major modes. Since $G_{\phi}$ already aims to cover all modes during the cooperative initialization stage, it is less likely to dropping major modes it already covers at the second stage. As to the discriminator or the descriptor $D_{\theta}$, in the stage of cooperative initialization, the output of the descriptor $D_{\theta}$ is a score representing negative energy. Real data typically receives higher scores (i.e., lower energy) from descriptor $D_{\theta}$. Similarly, in the adversarial finalization stage, the discriminator $D_{\theta}$ assigns larger probabilities to real data. Thus, both the descriptor and the discriminator can be viewed as classifiers with a shared objective. This allows us to initialize the discriminator with the descriptor.

\section{Related Work}\label{sec:work}
The following themes are closely related to our work, and we will briefly review each of them and explain their connection to our work. 
\par \vspace{-1mm} 
\paragraph{Regularization Techniques for GANs:} This line of research is based on both theoretical investigations and empirical studies on the convergence properties of GANs, in which regularization is used to ensure a good local equilibrium with new model assumptions. Various research efforts have been made in this direction, e.g., adding loss penalty~\cite{gulrajani2017improved,mescheder2018training}, weight regularization~\cite{miyato2018spectral,brock2018large} and implementing a discriminator bottleneck~\cite{zhao2020feature}. We can interpret the CoopInit as a special regularization technique, which only takes effect at the early stage of the learning process, to enforce the model to cover most of the modes in the data~distribution. 

\par \vspace{-1mm} 
\paragraph{Link MLE to GAN: }
The most successful works in linking MLE to GAN exist in the applications of GAN-based text generation~\cite{yu2017seqgan,nie2018relgan}. To mitigate the gradient estimation difficulty and mode collapse issues on discrete data, they apply large amount of MLE pretraining and limited adversarial fine-tuning. 
However, our CoopInit simultaneously trains both networks as a whole in an MLE-based cooperative manner using very limited time, whereas MLE pretraining used in~\citet{yu2017seqgan} trains them separately for most of the time. Besides, another work Flow-GAN~\cite{grover2018flow} uses a normalizing flow \cite{KingmaD18} as the generator to build a GAN. But, the expressive power of a normalizing flow is limited due to its restrictive network design.  \citet{zhao2020bridging} explore unifying the advantages of MLE and adversarial learning via $\alpha$-divergence but only trains the generator by MLE. Our method seamlessly bridges the MLE and GAN by the energy-based cooperative learning. 

\par \vspace{-1mm} 
\paragraph{Link EBM to GAN: }
Several works have investigated the relationship between EBMs and GANs~\cite{finn2016connection,che2020your}. Among these, DDLS~\cite{che2020your} is the most relevant, as it considers the discriminator as an energy function and employs MCMC in the latent space to generate refined samples. But, our CoopInit differs from DDLS in that we jointly train an EBM and a generator before GAN training, whereas DDLS only refines samples via MCMC after GAN training, without explicitly training~an~EBM. 

\section{Experiments}\label{sec:experiments}

In this section, we extensively evaluate the effectiveness of our proposed initialization strategy, CoopInit, for GANs. We begin by testing our method on image generation and unpaired image-to-image translation tasks, comparing our framework to state-of-the-art models.
Then we test the CoopInit across different loss functions, hyper-parameter settings, network architectures, model scales, and limited datasets. Finally, we present an ablation study to understand how the CoopInit works. All experiments were conducted on four Nvidia Titan Xp (12GB) GPUs and Google Colab.

\subsection{Experimental Setup}
\paragraph{Base Model}
In terms of performance, StyleGAN2 is currently the most attractive GAN model that can achieve state-of-the-art results on a variety of image synthesis tasks, such as image generation~\cite{karras2020training,zhao2020differentiable}, image translation~\cite{richardson2020encoding,zhao2020unpaired} and image manipulation~\cite{abdal2019image2stylegan}. StyleGAN2-ADA~\cite{karras2020training} is a specifically tuned GAN with techniques such as shallow mapping, disable style mixing regularization~\cite{karras2019style}, path length regularization, and residual connections in the discriminator. This model currently achieves state-of-the-art results on CIFAR-10 ~\cite{Krizhevsky09learningmultiple} image generation among all GANs. BigGAN~\cite{brock2018large} that was designed for generating  high-resolution and high-fidelity images is also considered.

\begin{figure*}[!ht]
    \centering
    \begin{tabular}{c|c}
        \includegraphics[width=0.48\linewidth]{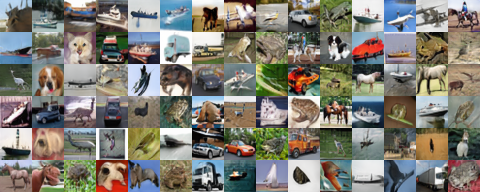} &  
        \includegraphics[width=0.48\linewidth]{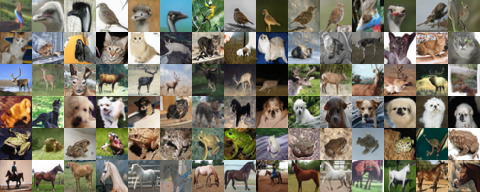} \\
    \end{tabular}
    
    \caption{Generated examples by the StyleGAN2-CoopInit-ADA models trained on the CIFAR10 dataset. (Left: Unconditional generation. Right: Conditional generation.) }\label{fig:cifar10_cropped}
\end{figure*}

\paragraph{Datasets} We evaluate the performance of image generation on four widely used datasets listed below: 

(i) CIFAR-10~\cite{Krizhevsky09learningmultiple}: This dataset consists of  60K $32\times32$ images in 10 evenly distributed classes, including 50K training images and 10K testing images. 

(ii) CIFAR-100~\cite{Krizhevsky09learningmultiple}: This dataset has 60K images in 100 classes, with 600 images per class for training. 

(iii) ImageNet~\cite{RussakovskyDSKS15}: To balance the computational budget, we use a down-sampled version of ImageNet that consists of $32\times32$ images. ImageNet contains over 10 million natural images of 1,000 classes.

(iv) FFHQ~\cite{karras2019style}: This dataset consists of 70K high-quality and diverse human facial images. We choose to use a down-sampled version of the data with a resolution of $256\times256$.

\paragraph{Metric}  Fr{\`e}chet Inception Distance (FID)~\cite{heusel2017gans} is a widely used metric for evaluating the quality of generated images. It computes the distance between the Inception feature vectors for real and generated images.
It is also consistent with increasing disturbances and human judgment. A low FID indicates that the model
can create high-quality images. 
We adopt the commonly used $50$K-FID, which generates 50K examples to evaluate image generation quality, as in most contemporary GAN works.


\subsection{Image Generation}
\subsubsection{Evaluation on CIFAR-10 Dataset}
We compare the proposed approach with state-of-the-art models on CIFAR-10 generation, and the results are shown in Table~\ref{tab:sota-cmp}. It is worth noting that when we disable the adaptive discriminator augmentation (ADA) in the base model StyleGAN2-ADA, the CoopInit can greatly reduce the FID from 6.40 to 4.34, even without $R_1$ regularization~\cite{mescheder2018training} (i.e., we set the hyperparameter of $R_1$ regularization $\gamma=0$). This is currently the best FID achieved by GANs on CIFAR-10 without using ADA. 
We further find that increasing the network depth hurts performance. When we double the width, the performance of  StyleGAN2-CoopInit with tuning is on par with that of the NCSN++cont.~\cite{song2021score} and achieves a new state-of-the-art result. 
We report the best FID of the generated images and evaluate the Inception Score (IS).
Figure~\ref{fig:cifar10_cropped} shows some generated examples.

\begin{table}[!ht]
\centering
    \begin{tabular}{l|c|c}
     \hline
        Models & FID$\downarrow$ & IS$\uparrow$ \\ \hline
        \multicolumn{3}{l}{\textbf{Conditional}} \\ \hline
        BigGAN &  &  \\
        \citep{brock2018large} & 14.73 & 9.22 \\
        MultiHinge~\cite{kavalerov2019cgans} & 6.40 & 9.58 \\
        FQ-GAN~\cite{zhao2020feature} & 5.59 & 8.48 \\
        \cdashline{1-3}
        BigGAN + CoopInit (ours) & 6.95 & 9.35\\
        StyleGAN2 {\scriptsize w/ ADA}~\cite{karras2020training} &  2.42 & 10.14   \\
        + CoopInit + tuning (ours) &  {\bf 2.20} &  {\bf 10.20} \\ \hline
        \multicolumn{3}{l}{\textbf{Unconditional}}\\ \hline
        CoopNets~\cite{xie2018cooperativePAMI} & 33.61 & - \\
        CoopVAEBM~\cite{XieZL21} & 36.20 & - \\
        CoopFlow~\cite{XieZLL22} & 15.80 & - \\
        CF-EBM~\cite{zhao2021learning} & 16.71 & - \\
        ProGAN~\cite{karras2017progressive} & 15.52 & 8.56 \\
        NCSNv2~\cite{song2020improved} & 10.87 & 8.40 \\
        CAS~\cite{jolicoeur-martineau2021adversarial} & 3.65 & - \\
        DDPM~\cite{ho2020denoising} & 3.17 & 9.46 \\
        StyleGAN2-ADA~\cite{karras2020training} & 2.92 & 9.83 \\
        NCSN++cont.~\cite{song2021score} & {\bf 2.20} & 9.89\\ \cdashline{1-3}
        StyleGAN2 {\scriptsize w/o ADA} ($\gamma=0.01$)   & 6.40 & 9.55 \\
        + CoopInit (ours) ($\gamma=0.00$) & 4.34 &  9.69 \\ 
        StyleGAN2 {\scriptsize w/ ADA}~\cite{karras2020training} &  2.92 &   9.83 \\
        + CoopInit (ours) &  2.82  &   9.88\\
        + CoopInit + tuning (ours) &  {\bf 2.55} &  {\bf 9.94} \\
        \hline
    \end{tabular}
    
    \captionof{table}{FID and Inception score (IS) comparison on conditional and unconditional CIFAR-10 image generation.}
    \label{tab:sota-cmp}
\end{table}

\subsubsection{Evaluation on FFHQ Dataset}

\begin{table}[!htbp]
    \centering
  
    \begin{tabular}{l|c} 
    \hline
        Models & FID $\downarrow$ \\ \hline
        BigGAN~\citep{brock2018large} & 11.48 \\
        U-Net GAN &  \\
        ~\cite{schonfeld2020u} & 7.48 \\ \cdashline{1-2}        
        StyleGAN2 & 3.84 \\         
        + CoopInit \;(ours) & {\bf 3.61} \\       
        \hline
    \end{tabular}
    \caption{FID comparison on FFHQ $256\times 256$.}
    \label{tab:other}
\end{table}

\begin{figure*}[!htbp]
\centering
\includegraphics[width=0.99\linewidth]{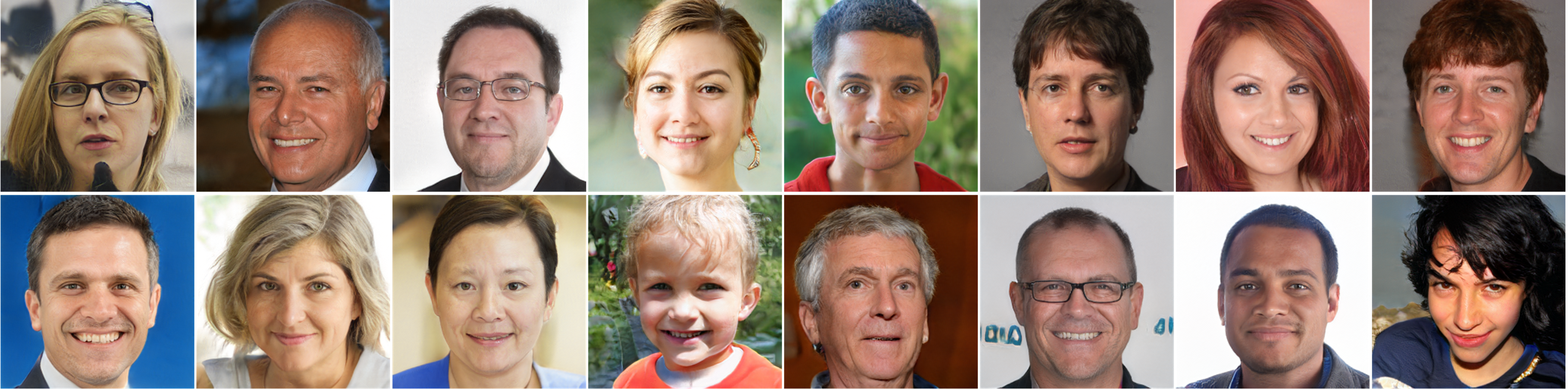}
\caption{Qualitative results of FFHQ $256 \times 256$ image generation.}
\label{fig:FFHQ}
\end{figure*}

Compared to CIFAR-10 and CIFAR-100 datasets, the image distribution of FFHQ dataset is more concentrated but less diverse. 
The CoopInit method can consistently outperform the baseline, as shown in Table~\ref{tab:other}. Qualitative results are presented in Figure~\ref{fig:FFHQ}.

\subsubsection{Evaluation on ImageNet Dataset}
In our previous study, we show that CoopInit can significantly improve the performance of GANs in various scenarios. To further evaluate its effectiveness, we conduct a study on a more complex dataset, ImageNet. The results in Table~\ref{tab:imagenet} indicate that although CoopInit performs better on unconditional generation, its performance on conditional generation is only comparable to the baseline. We suspect that this is because the label information can alleviate the mode collapse issue to some extent, which aligns with the objective of CoopInit.

\begin{table}[!htbp]
    \centering
    \begin{tabular}{l|c}

    \hline
    Models & FID$\downarrow$  \\ \hline
        \multicolumn{2}{l}{\textbf{ImageNet} $(32\times32)$} \\ \hline
        PixelCNN~\cite{van2016conditional} & 33.27\\
        PixelIQN~\cite{ostrovski2018autoregressive}  & 22.99\\
        IGEBM~\cite{du2019implicit}  & 14.31 \\ \cdashline{1-2}
        StyleGAN2 {w/o} labels   & 6.87 \\ 
        + CoopInit (ours) & {\bf 5.84} \\
        StyleGAN2 {w/} labels   & 3.87 \\ 
        + CoopInit (ours) & 3.84 \\
        \hline
        \multicolumn{2}{l}{\textbf{ImageNet} $(64\times64)$} \\ \hline
        BigGAN {w/} labels &  \\
        ~\citep{brock2018large}& 10.55\\
        +CoopInit (ours) & 10.63 \\
        \hline
    \end{tabular}
    \caption{FID comparison on ImageNet dataset.}
    \label{tab:imagenet}
\end{table}

\subsection{Unpaired One-sided Image Translation}
\begin{figure}[h]
\centering
\includegraphics[width=0.45\textwidth]{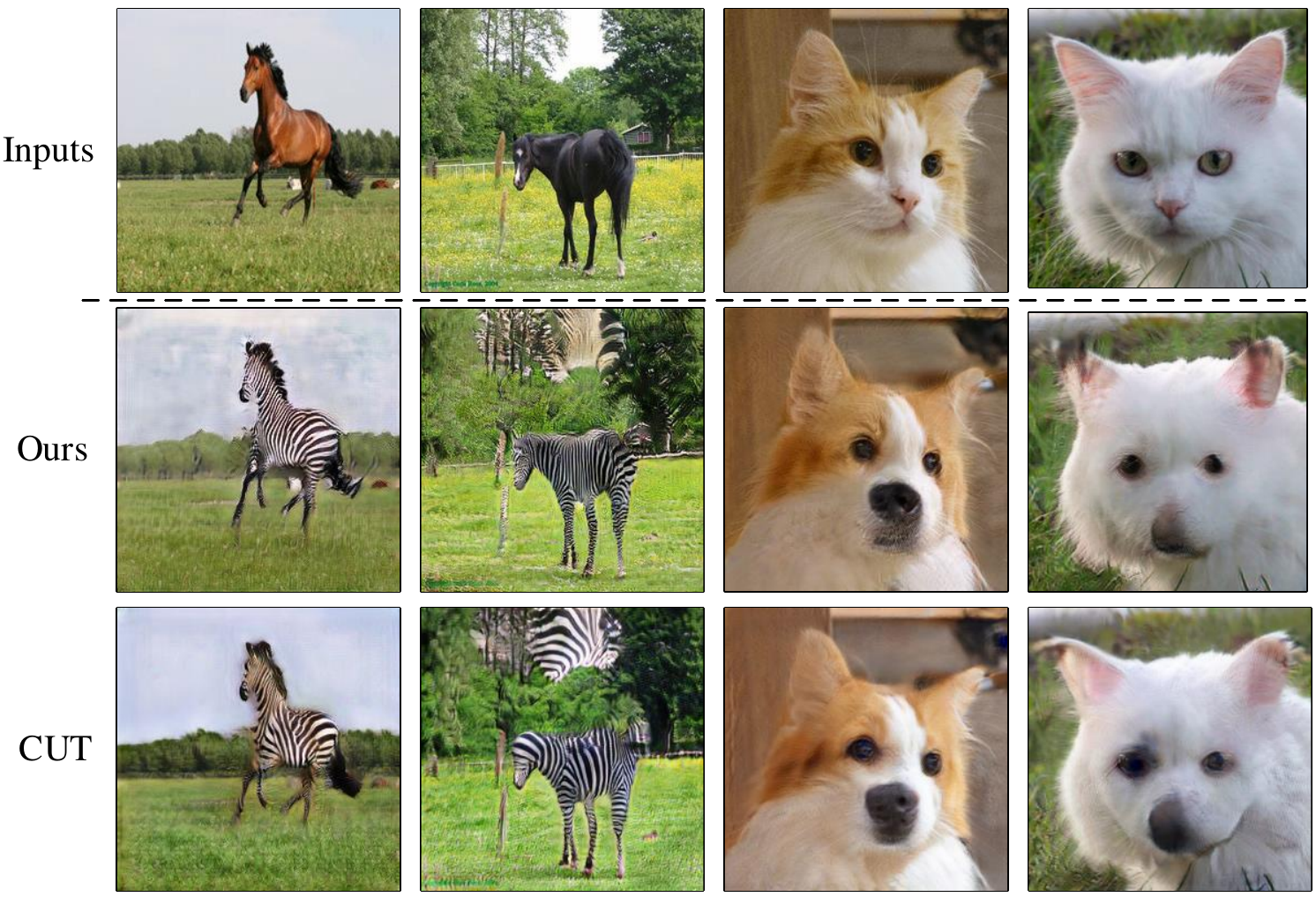}
\caption{Comparison of qualitative results for one-sided unpaired image-to-image translation using a baseline CUT and our method CUT+CoopInit.}
\label{fig:one-sided}
\end{figure}

The proposed CoopInit is also tested in the context of adversarial image-to-image translation. We evaluate our approach on the recently proposed approach CUT~\cite{park2020contrastive}, which enables one-sided image-to-image translation using patch-wise contrastive learning and adversarial learning for content preservation and style transfer. The results, both quantitative and qualitative, shown in Table~\ref{tab:i2i-cmp} and Figure~\ref{fig:one-sided}, outperform the baselines. We observe an improvement in the performance of CUT when CoopInit is employed. The baseline method CF-EBM~\cite{zhao2021learning} is an energy-based model that uses short-run Langevin dynamics as a flow-like generator to transform images from the source domain to the target domain. We encountered difficulties when applying CF-EBM to the Horse$\Rightarrow$Zebra task, and we suspect that this may be due to misalignment between the source and target datasets. Additionally, it is worth noting that in the cooperative initialization stage, our generator performs a direct transformation of the source domain images to the target domain. The output is then fed into the Langevin dynamics of the descriptor  for a few steps of revision. Compared to CF-EBM, CoopInit employs a top-down generator to amortize the computationally expensive MCMC process.

\begin{table}[ht!]
\centering
\begin{tabular}{l|cc}
\hline
\multirow{2}{*}{Models} & \multicolumn{2}{c}{FID$\downarrow$} \\ 
& C$\Rightarrow$D & H$\Rightarrow$Z\\ \hline
Distance~\cite{benaim2017one} & 155.3 & 72.0  \\ 
SelfDistance~\cite{benaim2017one} & 144.4 & 80.8 \\
GCGAN~\cite{fu2019geometry} & 96.6 & 86.7  \\ 
CF-EBM~\cite{zhao2021learning} & \textbf{55.1} & - \\
\cdashline{1-3}
CUT~\cite{park2020contrastive} & 76.2 & 45.5  \\
+ CoopInit (ours) & {61.3} & {\bf 38.7}\\
\hline
\end{tabular}
\caption{Comparison on one-sided unpaired image-to-image translation. (C$\Rightarrow$D: Cat$\Rightarrow$Dog. H$\Rightarrow$Z: Horse$\Rightarrow$Zebra)}
\label{tab:i2i-cmp}
\end{table}

\subsection{Model Analysis}
To investigate the impact of adversarial loss functions, hyperparameters and neural architecture designs, we test the CoopInit on CIFAR-10 dataset for image generation. Following~\citet{zhao2020differentiable}, we halve the number of channels of feature maps at higher resolution layers (i.e., $16\times16$ and above) to enable faster computation. We further apply the non-saturating loss, set the learning rate to $0.0025$, and use the original connection unless specified otherwise, following the approach of~\citet{karras2020training}. To ensure fair and consistent comparisons, we temporarily disable the lazy mode of $R_1$ regularization. This is because the lazy mode leads to a different optimization, which requires a decrease in the learning rate and hyperparameters in the Adam optimizer~\cite{karras2020analyzing}. We use 100M real images for each run with data augmentation and 25M without.

\subsubsection{Impact of Loss Functions}

We conduct an investigation into the impact of different adversarial loss variants on training GANs, including Hinge loss (Hinge), non-saturating loss (NS), and Wasserstein distance with gradient penalty (WAS-GP). After extensive hyper-parameter tuning, we select the best learning rate and report the FIDs in Table~\ref{tab:loss-cmp}. To ensure a fair comparison, all tests share the same architecture, and each column uses the same optimizers. As shown in Table~\ref{tab:loss-cmp}, CoopInit consistently yields lower FIDs on the three loss variants, with the most significant improvements observed on Hinge and NS losses. We also find that the default NS loss with $R_1$ regularization is the most appropriate loss function to train StyleGAN2, but this is no longer the case when CoopInit is applied. Interestingly, CoopInit with StyleGAN2-Hinge is found to yield a much better FID compared with StyleGAN2-NS with $R_1$ regularization.

\begin{table}[!htbp]    
\centering
  
    \begin{tabular}{c|ccc}
    \hline
        Methods & NS & Hinge & WAS-GP \\ \hline
        StyleGAN2 & $13.95$ $(8.71^\ast)$ & $11.64$ & $13.21$ \\ 
        + CoopInit (ours) & ${\bf5.85}$ & ${\bf5.09}$ & ${\bf 11.83 }$ \\ \hline
    \end{tabular}
    \captionof{table}{
    CoopInit improves StyleGAN2 with different variants of adversarial loss. The sign $^\ast$ indicates a performance obtained using $R_1$ regularization with $\gamma=0.01$.}
    \label{tab:loss-cmp}
\end{table}

\subsubsection{Comparison with Other Regularization Techniques}

Table~\ref{tab:reg} presents a comparison of our proposed CoopInit technique with other well-known GAN regularization methods. The base model is a full-sized StyleGAN2-NS without using $R_1$ regularization. For the WGAN-GP method, we replace the non-saturating loss with the Wasserstein distance with gradient penalty. As shown in the table, our proposed CoopInit outperforms the baselines. Moreover, CoopInit has the added benefit of computational efficiency, especially when considering the cost of gradient penalty.

\begin{table}[!ht]
    \centering
    
    \begin{tabular}{l|c}
        \hline
        Methods & FID $\downarrow$\\ \hline
        Base model  & 15.8 \\ 
        + $R_1$ regularization\\ ~\cite{mescheder2018training} & 6.40 \\ 
        + $R_1$ + Spectral Norm~\cite{miyato2018spectral}  & 6.98 \\
        + $R_1$ + zCR~\cite{zhao2020improved} & 5.71 \\ 
        + CoopInit (ours) & {\bf 4.34} \\
        WGAN-GP~\cite{gulrajani2017improved} & 12.33 \\
        \hline
    \end{tabular}
    \captionof{table}{Comparing CoopInit with other regularization techniques in the base model StyleGAN2-NS. }
    \label{tab:reg}
\end{table}

\subsubsection{Impact of Hyperparameters}
In this section, we evaluate the impact of some hyperparameters, including the learning rate $lr$ and the strength of the $R_1$ regularization $\gamma$, on the proposed learning algorithm. 

\begin{table}[!ht]
         \centering
        \captionsetup{type=figure}
         \includegraphics[width=0.98\linewidth]{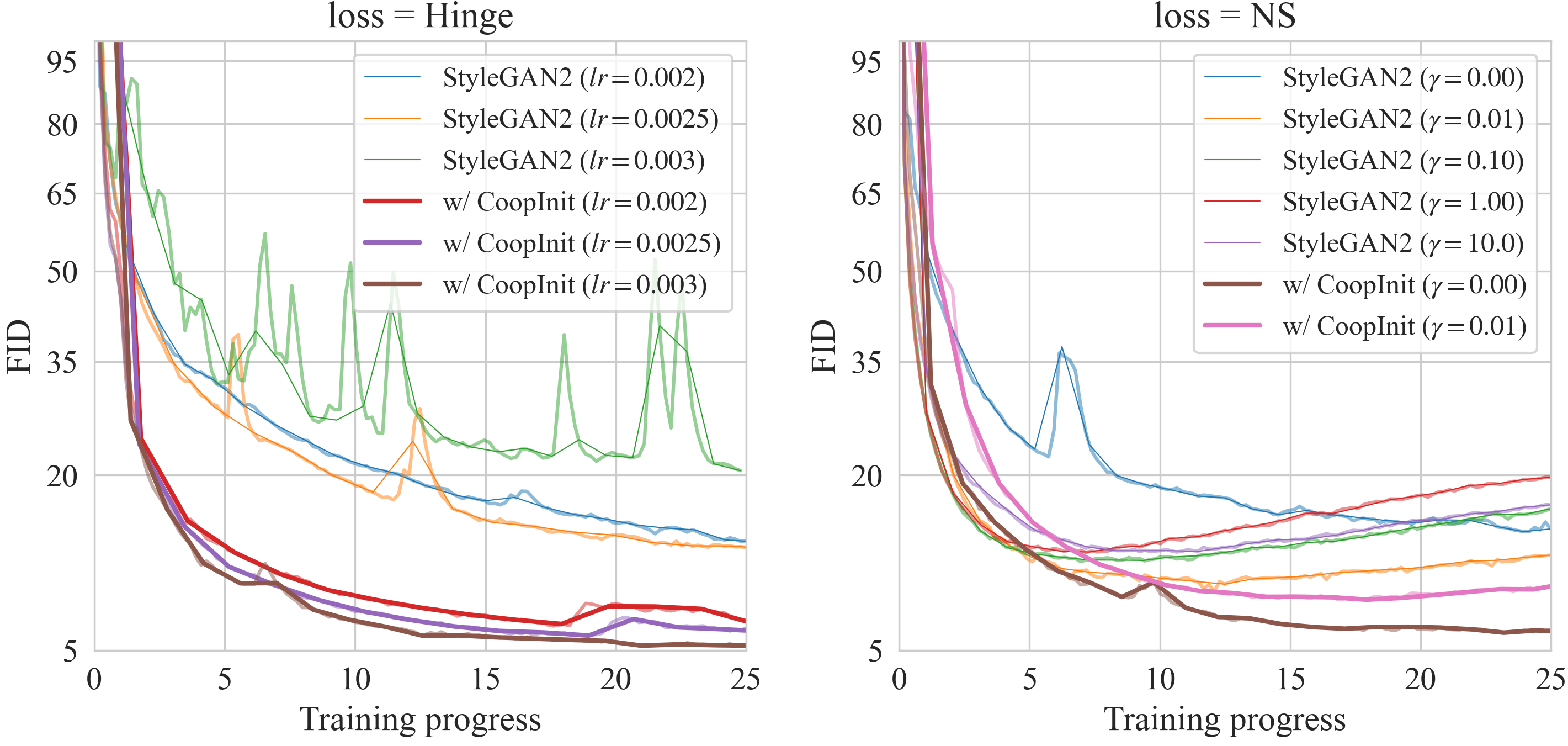}
          \captionof{figure}{Learning curves of different GAN variants.}
    \label{fig:hyper-cmp}
\end{table}

\begin{table}[!ht]
\centering
    \begin{tabular}{c|c|ccc}
    \hline
        \multirow{2}{*}{Methods} & \multirow{2}{*}{$\gamma$} & \multicolumn{3}{c}{$ lr \times10^{-3}$} \\
        & & 2.0 & 2.5 & 3.0 \\ \hline
        StyleGAN2 &  \multirow{2}{*}{0.00} & 13.21 & 13.95 & 14.94 \\ 
        + CoopInit (ours) & & 6.16 & 6.07 & {\bf 5.85} \\ \hline
        StyleGAN2 & \multirow{2}{*}{0.01} & 9.28 & 8.95 & 8.71 \\ 
        + CoopInit (ours) & & 8.87   &  8.29 & {\bf 7.58}  \\ 
        \hline
    \end{tabular}
    \captionof{table}{CoopInit improves StyleGAN2-NS across different learning rates ($lr$) and two $R_1$ regularization~settings.}
    \label{tab:ns-lr}
\end{table}
\textit{Learning Rate.}
We conduct two sets of experiments with non-saturating (NS) loss and Hinge loss, respectively, to study how our CoopInit technique behaves when the learning rate varies. The results are shown in Figure~\ref{fig:hyper-cmp} (left) and Table~\ref{tab:ns-lr}. As shown in Figure~\ref{fig:hyper-cmp} (Left), StyleGAN2-Hinge benefits  greatly from the CoopInit technique across all different learning rates. In particular, when we increase the learning rate $lr$ to 0.003, CoopInit can eliminate the acute oscillation of the original StyleGAN2-Hinge and drive the model to reach the fastest convergence rate among all learning rate settings. This verifies the effectiveness of CoopInit and the importance of a good initial point for GAN training. Table~\ref{tab:ns-lr} also confirms the results under different $R_1$ regularization hyperparameters $\gamma$ with  NS loss.

\textit{$R_1$ Regularization Strength.} $R_1$ regularization is a critical technique to stabilize StyleGAN2-NS training and helps to reach a local equilibrium faster~\cite{mescheder2018training}. In the right panel of Figure~\ref{fig:hyper-cmp}, we plot learning curves for models using various values of $\gamma$, which is a hyperparameter in $R_1$ regularization and represents the strength. We observe that StyleGAN2-NS is very sensitive to the regularization strength, and the performance deteriorates after some iterations. We find that $\gamma=0.01$ works best, which is consistent with ~\citet{karras2020training}. In contrast, the minimum FID and the most stable learning curve can be obtained when we replace $R_1$ regularization by CoopInit in training StyleGAN2-NS. Results in Table~\ref{tab:ns-lr} demonstrates that CoopInit works best without using $R_1$ regularization. 

\subsubsection{Impact of Neural Architectures}
The design of neural architecture is always one of the most critical factors to improve GANs, e.g., BigGAN~\cite{brock2018large} and StyleGAN~\cite{karras2019style}. We use  StyleGAN2 as the base model. To study the compatibility between CoopInit and different network architectures, we investigate the effects of using original (Orig), residual (Res), and skip (Skip)  convolutional layer connections, as well as a label conditional layer (Cond) in the discriminator. Table~\ref{tab:arch-cmp} shows that CoopInit significantly improves the results of the base model for all three architectures. It is worth noting that, for the base model, the original (Orig) architecture obtains the lowest FID, which is consistent with the findings in \citet{karras2020training}. However, when the CoopInit is applied to the base model, the Skip architecture proves to be the most effective one. We find that the conditional information can decrease the performance when CoopInit is applied, but the result still outperforms the base model. We suspect that the projection discriminator~\cite{miyato2018cgans} may not the most effective method for CoopInit to learn conditional~information. 
\begin{table}[!htbp]
    \centering
    \begin{tabular}{c|ccc|c}
    \hline
        Methods & Res & Skip & Orig & + Cond \\ \hline
        StyleGAN2 ${ (\gamma=.01)}$ & 10.85 & 9.39 & 8.95 & 7.94 \\
        + CoopInit ${ (\gamma=.00)}$ & {\bf 6.01} & {\bf 5.46} & {\bf 6.07} & {\bf 6.67} \\
        \hline
    \end{tabular}
    \caption{Impact of discriminator architectures on CoopInit.}
    \label{tab:arch-cmp}
\end{table}

\begin{figure}[!ht]
    \centering
    \includegraphics[width=0.99\linewidth]{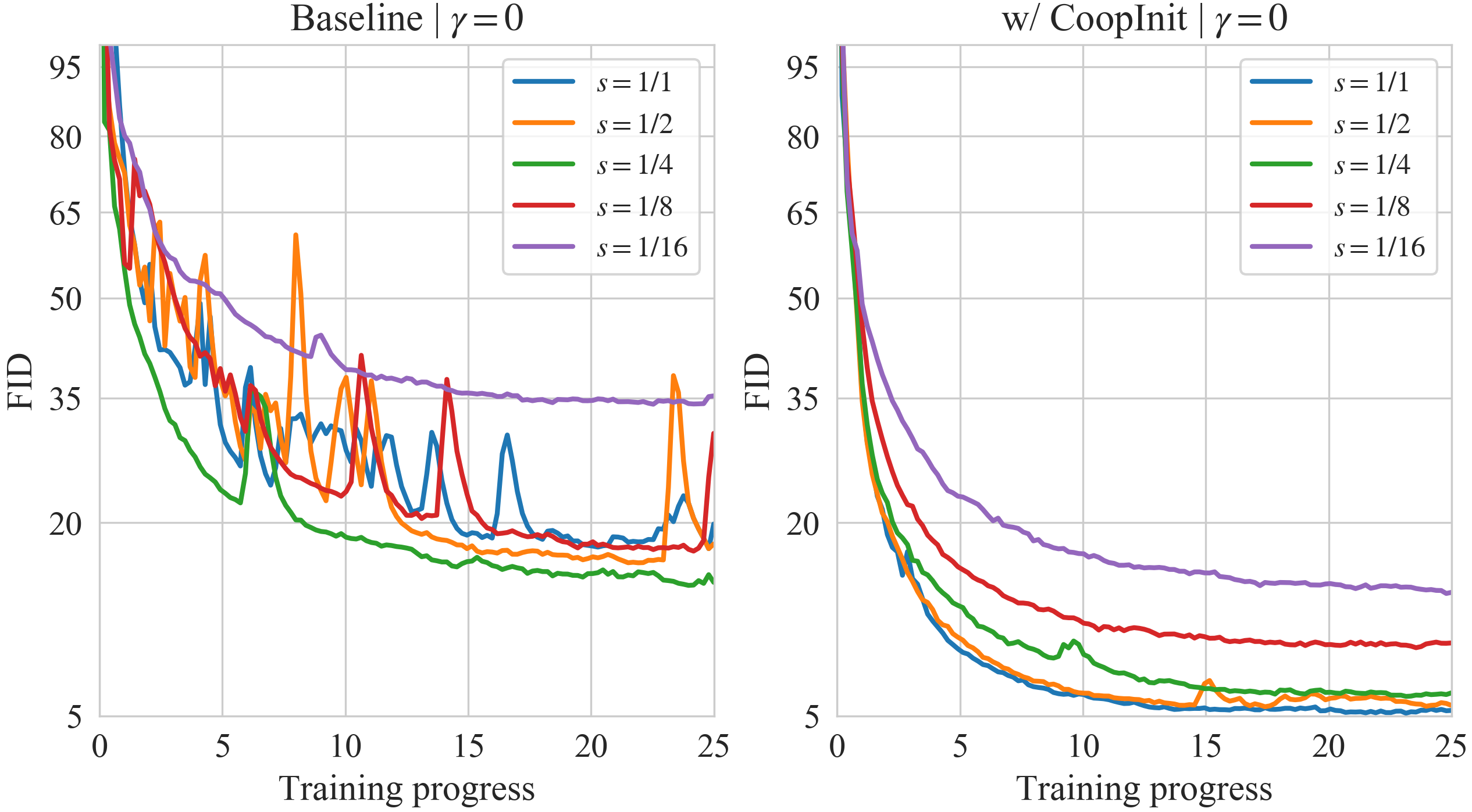}
    \caption{Comparison of learning curves of models with different model capacities. (Left: StyleGAN2 without $R_1$ regularization. Right: unregularized StyleGAN2 with CoopInit). }
    \label{fig:model-scale-cmp}
\end{figure}
\subsubsection{Model Capacity}  To examine the impact of the overall model capacity, we vary the multiplier $M\cdot s$ of the network architecture in order to control the sizes of both the generator $G$ and the descriptor (or discriminator) $D$. In this context, $M$ refers to the predefined multiplier ($M=2^{13}$) in~\citet{karras2020training} while $s$ is a scaling factor that allows for the adjustment of model capacity. The base model used in this section corresponds to $s=1/4$. The left panel of Figure~\ref{fig:model-scale-cmp} shows that unregularized StyleGAN2 (i.e., $\gamma=0$) is highly sensitive to changes in model capacity, with oscillations observed in learning curves for different model sizes. This suggests that manual hyperparameter tuning is necessary for each model scale. We observe that applying lazy regularization and $R_1$ regularization with $\gamma=0.01$ reduces the FID of the base model from 17.56 to 6.40 for $s=1/1$. In contrast, CoopInit appears to be less dependent on manual parameter tuning and consistently delivers significant improvements. 

\subsubsection{Unbalanced $D$ and $G$}
Most GANs are designed to balance the capacities of the generator $G$ and the discriminator $D$. However, the reason for this balance is not immediately apparent. Functionally, the generator acts as an ancestral sampler while the discriminator acts as a classifier. Therefore, there seems to be no intuitive reason to strive for such a balance between $D$ and $G$ as they play entirely different roles. In our study, we test the effectiveness of the CoopInit in unbalanced GAN scenarios. We fix the generator scale factor $s=1/4$ and vary the discriminator's scale factor. As shown in Table~\ref{tab:unbalanced-cmp}, CoopInit significantly improves the baseline across all different unbalanced settings. The results verifies the effectiveness of CoopInit in mitigating the mode collapse and instability issues in adversarial learning.

\begin{table}[!ht]
\centering
        \begin{tabular}{c|ccccc}
       \hline
          \multirow{2}{*}{Methods} & \multicolumn{5}{c}{Discriminator scale ($s$) ($\gamma=0$)} \\ 
            & $1/16$ & $1/8$ & ${1/4}$ & $1/2$ & $1/1$ \\ \hline
            StyleGAN2 & 35.52 & 22.14 & {13.95} & 14.02 & 13.15 \\ 
            + CoopInit & {\bf 16.45} & {\bf 9.00}  & {\bf 6.07} & {\bf 5.32} & {\bf 5.26} \\
            \hline
        \end{tabular}
        \captionof{table}{CoopInit improves unbalanced GANs. }
        \label{tab:unbalanced-cmp}
\end{table}

\subsubsection{Limited Data}
We then test our method on limited amounts of training data. We conduct experiments on CIFAR-10 and CIFAR-100 generation with limited data. As shown in Tables~\ref{tab:limited-cmp-da} and ~\ref{tab:limited-cmp-da2}, CoopInit significantly outperforms the baseline across all scenarios. In particular, the improvements are more significant when the ADA~\cite{karras2020training} is not used. The ADA can only be enabled when the adversarial loss is the NS loss with appropriate $R_1$ regularization, as the augmentation strength is determined by the output of the discriminator. Meanwhile, lazy regularization will also affect the final performance with ADA. Therefore, we disable the ADA in the CoopInit phase and only enable the lazy regularization for the results in Table~\ref{tab:sota-cmp}. We follow all settings according to \citet{karras2020training} but change the learning rate to 0.003, which shows better performances. We adopt the ADA technique introduced in StyleGAN2-ADA~\cite{karras2020training}. As suggested by~\citet{karras2020training}, we use the whole training dataset as the reference distribution and generate 50K examples to compute FID. The reported FIDs are averaged from the best FIDs in three different runs. %

\begin{table}[!ht]
    \centering
    \begin{tabular}{lcc|ccc}
    \hline
        \multirow{2}[2]{*}{Methods} & \multirow{2}[2]{*}{$\gamma$} & \multirow{2}[2]{*}{ADA} &  \multicolumn{3}{c}{Number of training data} \\ 
         & & & 50K & 20K & 10K  \\ \hline
          StyleGAN2 & $0.01$ & \xmark   &  8.71  & 18.80 & 27.68    \\
         + CoopInit & $0.00$ & \xmark &{\bf 5.12} &  {\bf 14.36} & {\bf 20.95} \\ \hline
         StyleGAN2 & $0.01$ & \cmark    & 3.62 & 4.60 & 6.92 \\
        + CoopInit & $0.01$ & \cmark & {\bf 3.59}  & {\bf 4.49} & {\bf 6.58} \\ \hline
    \end{tabular}
    \caption{FID results of models trained on limited amounts of training data from CIFAR-10 dataset.}
    \label{tab:limited-cmp-da}
\end{table}

\begin{table}[!ht]
    \centering
    \begin{tabular}{lcc|ccc}
    \hline
        \multirow{2}[2]{*}{Methods} & \multirow{2}[2]{*}{$\gamma$} & \multirow{2}[2]{*}{ADA} &   \multicolumn{3}{c}{Number of training data}\\ 
         & & & 50K & 20K & 10K \\ \hline
          StyleGAN2 & $0.01$ & \xmark   &  11.40 & 24.24 & 34.53  \\
         + CoopInit & $0.00$ & \xmark & {\bf 8.10} & {\bf 19.20}  & {\bf 29.28}\\ \hline
         StyleGAN2 & $0.01$ & \cmark    & 5.00 & 6.82 & 9.63 \\
         + CoopInit & $0.01$ & \cmark &  5.10  & {\bf 6.45} & {\bf 8.80} \\ \hline
    \end{tabular}
    \caption{FID results of models trained on limited amounts of training data from CIFAR-100 dataset. }
    \label{tab:limited-cmp-da2}
\end{table}

\subsubsection{Duration of CoopInit}  
\begin{figure}[!ht]
    \centering
    \includegraphics[width=0.85\linewidth,height=0.7\linewidth, trim={0 0 0 0.5cm},clip]{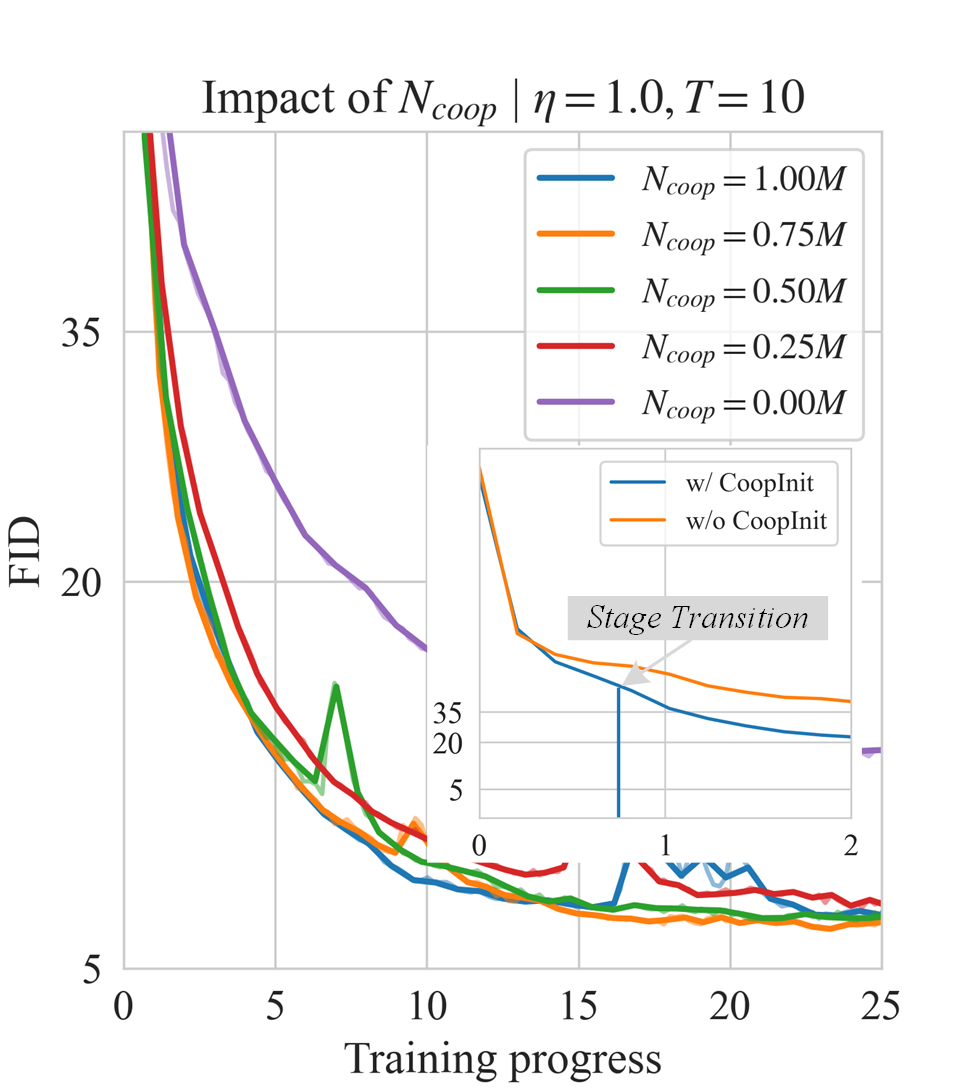}
    \caption{Impact of the duration of cooperative initialization. $N_{\text{coop}}$ is the number of training data consumed by the algorithm at the cooperative initialization stage. The embedded plot illustrates smooth stage transitions.}
    \label{fig:ncoop-cmp}
\end{figure}

Notation $N_{\text{coop}}$ represents the number of training examples consumed in the cooperative initialization stage. In general, a larger value of $N_{\text{coop}}$ corresponds to a longer duration of the CoopInit process, resulting in a greater number of modes being covered by the model. However, this comes at the cost of an increase in the training time due to the additional MCMC computation required in the CoopInit stage. 
In Figure~\ref{fig:ncoop-cmp}, we vary $N_{\text{coop}}$ to compare the performance and find that CoopInit can significantly improve both convergence rate and final sample quality for GANs. Meanwhile, we also observe that there is no  straightforward positive correlation between  the duration of cooperative learning $N_{\text{coop}}$ and the image generation quality quantified by FID, and that $N_{\text{coop}}=0.75$M appears to be the optimal choice under the current experimental settings. We even observe an instability issue in a model with a much larger value of $N_{\text{coop}}$. The embedded plot in Figure~\ref{fig:ncoop-cmp} shows a smooth stage transition between cooperative learning and adversarial learning in a GAN training using CoopInit.

\subsubsection{Langevin Step Size $\eta$ and Number of Langevin Steps $T$} 

The hyperparameters $\eta$ and $T$ are closely related to the cooperative learning shown in Algorithm~\ref{alg:coopnet}. Generally, smaller values of $\eta$ and $T$ may not be sufficient for learning the model to capture the target distribution, while larger values of $\eta$ and $T$ can result in instability issues and high computational costs, respectively. In this experiment, we set $N_{\text{coop}}$ to be 0.75M. As shown in Table~\ref{tab:coop-t}, increasing the step size leads to better performance. Thus, we recommend using $\eta=1.0$, as it works well across different datasets and models. Moreover, as seen in Table~\ref{tab:coop-t}, the impact of the number of Langevin steps, $T$, on CIFAR-10 generation is minimal. However, a longer chain of Langevin dynamics is more appropriate when the dataset is more complicated and the resolution is higher. By default, we use $\eta=1.0$ and~$T=10$.

\begin{table}[!ht]
    \centering
    \begin{tabular}{c|ccc|ccc}
        \hline
        &\multicolumn{3}{c|}{$\eta$ ($T$=10)} & \multicolumn{3}{c}{$T$ ($\eta$=1.0)} \\   
        & 0.5 & 1.0 & 5.0 & 5 & 10 & 15\\  \hline
        FID$\downarrow$ & 6.61 & \textbf{6.07} & 6.10 & \textbf{5.93} & 6.07 & 5.95\\  \hline
        \end{tabular}
        \caption{Impact of hyperparameters for Langevin dynamics. $\eta$: Langevin step size. $T$: Number of Langevin steps.}
        \label{tab:coop-t}
\end{table}

\section{Conclusion}\label{sec:conclusion}
To summarize, this paper establishes a new connection between cooperative learning and adversarial learning by proposing to adopt cooperative learning (i.e., CoopNets algorithm) to initialize GAN training. Our hybrid learning scheme, CoopInit, allows us to seamlessly integrate the strengths of both CoopNets and GAN, and it is compatible with various techniques for stabilizing and enhancing GANs. We demonstrate significantly improved performance across extensive experiments on a variety of datasets, including image generation and image-to-image translation. We also achieve a new state-of-the-art result for image generation on CIFAR-10 dataset. Future works can explore broader applications of the CoopInit technique, e.g., generative representation learning and  controllable image generation.

\clearpage
\newpage

\bibliography{aaai23}

\newpage

\appendix
\section*{Appendix}

\section{MLE is Equivalent to Minimizing Kullback-Leibler (KL) Divergence}
The cooperative learning seeks to maximize the likelihoods of both the generator and the descriptor. We here show that maximum likelihood estimation is equivalent to minimizing KL-divergence between the true data distribution $p_{\rm data}(x)$ and the model $p_{\theta}(x)$. Specifically,
\begin{eqnarray}
\begin{aligned}
&\mathsf{KL}(p_{\rm data}(x){\parallel}p_{\theta}(x)) \nonumber\\
=&\mathbb{E}_{x \sim p_{\rm data}(x)} \left[\log \frac{p_{\rm data}(x)}{p_{\theta}(x)}\right]\\
=&\mathbb{E}_{x \sim p_{\rm data}(x)} \left[\log p_{\rm data}(x) - \log p_{\theta}(x)\right]\\
=&\mathbb{E}_{x \sim p_{\rm data}(x)} [\log p_{\rm data}(x)] - \mathbb{E}_{x \sim p_{\rm data}(x)} [\log p_{\theta}(x)],
\end{aligned}
\end{eqnarray}
where the left term is the entropy of the data distribution that is not dependent on the model parameter $\theta$, thus we have 
\begin{eqnarray}
\begin{aligned}
&\arg \min_{\theta} \mathsf{KL}(p_{\rm data}(x){\parallel}p_{\theta}(x)) \\
=& \arg \max_{\theta} \mathbb{E}_{x \sim p_{\rm data}(x)} [\log p_{\theta}(x)].
\end{aligned}
\end{eqnarray}
Suppose we observe $n$ training examples $\{x_i\}_{i=1}^{n}\sim p_{\rm data}(x)$, according to the law of large number, if $n$ goes to infinity, 
\begin{eqnarray}
\begin{aligned}
\frac{1}{n} \sum_{i=1}^{n} \log p(x|\theta) =  \mathbb{E}_{x \sim p_{\rm data}(x)} [\log p_{\theta}(x)],
\end{aligned}
\end{eqnarray}
or in other words, if $n$ is large enough, the left term which is the data log-likelihood $\mathcal{L}(\theta|\{x_i\})$ can be used to approximate the right term, and therefore maximizing the data log-likelihood is equivalent to minimizing the KL-divergence between the data distribution and the model, i.e.,
\[
\arg \max_{\theta} \mathcal{L}(\theta) = \arg \min_{\theta} {\rm KL}(p_{\rm data}(x)|| p_{\theta}(x)).
\]

\section{Deriving the Gradient of the Likelihood of the Descriptor in CoopNets}
The descriptor in cooperative learning is an energy-based model or a Gibbs distribution, given by
\[
p_{\theta}(x)=\frac{1}{Z({\theta})}\exp[D_{\theta}(x)]
\]
over signal $x$, where $Z({\theta})=\int \exp[D_{\theta}(x)] dx$ is the intractable normalizing constant. The training of the descriptor seeks to find $\theta$ in the parameter space such that the parametric model $p_{\theta}(x)$ is able to get close to the data distribution $p_{\rm data}(x)$ in terms of KL-divergence or equivalently maximize the likelihood. To be specific,  
\begin{eqnarray}
\begin{aligned}
 &\nabla_\theta  \log p_{\theta}(x) = \nabla_\theta  D_{\theta}(x) -\nabla_\theta  \log Z(\theta),
\end{aligned}
\end{eqnarray}
where the term $\nabla_\theta  \log Z(\theta)$ can be rewritten as
\begin{eqnarray}
\begin{aligned}
\nabla_\theta  \log Z(\theta) & = \nabla_\theta  \log \int \exp[D_{\theta}(x)] dx\\
& =(\int \exp[D_{\theta}(x)] dx)^{-1} \nabla_\theta \int \exp[D_{\theta}(x)] dx\\
& =(\int \exp[D_{\theta}(x)] dx)^{-1}  \int \nabla_\theta \exp[D_{\theta}(x)] dx\\
& =\frac{1}{Z(\theta)}  \int  \exp[D_{\theta}(x)] \nabla_\theta D_{\theta}(x) dx\\
& =  \int  \frac{1}{Z(\theta)}\exp[D_{\theta}(x)] \nabla_\theta D_{\theta}(x) dx\\
& =  \int  p_{\theta}(x) \nabla_\theta D_{\theta}(x) dx\\
&= \mathbb{E}_{ p_{\theta}(x)}[\nabla_\theta D_{\theta}(x)]. \nonumber
\end{aligned}
\end{eqnarray}
Thus, 
\begin{eqnarray}
\begin{aligned}
 &\nabla_\theta  \log p_{\theta}(x)= \nabla_\theta  D_{\theta}(x) -\mathbb{E}_{p_{\theta}(x)}[\nabla_\theta D_{\theta}(x)],
\end{aligned}
\end{eqnarray}
and the gradient of the log-likelihood is 
\[
\nabla_\theta \mathcal{L}(\theta) = \mathbb{E}_{p_{\rm data}(x)} [\nabla_\theta D_{\theta}(x)] -\mathbb{E}_{p_{\theta}(x)}[\nabla_\theta D_{\theta}(x)].
 \]
 
\section{Loss Functions of GANs}

We present the definitions of different GAN loss variants, such as WAS, WAS-GP, and Hinge loss.

$(1)$ WAS:\\
WAS is a loss, based on Wasserstein distance, used in Wasserstein GAN. It is given by
\begin{equation} \label{eq:was}
  \underset{\phi}{\mbox{min}}\; \underset{\theta}{\mbox{max}} \; \mathbb{E}_{p_{\text{data}}(x)} [D_{\theta}(x)] - \mathbb{E}_{ p(z)} [D_{\theta}(G_{\phi}(z))].
\end{equation}
We find that our model using WAS loss fails to converge on the CIFAR-10 experiment.

$(2)$ WAS-GP:\\
WAS-GP adds an additional gradient penalty to the WAS loss, resulting in the following loss function
\begin{equation}\label{eq:was-gp}
    \begin{split} 
  \underset{\phi}{\mbox{min}}\; \underset{\theta}{\mbox{max}} \; &\mathbb{E}_{p_{\text{data}}(x)} [D_{\theta}(x)] - \mathbb{E}_{p(z)} [D_{\theta}(G_{\phi}(z))] \\
  & - \lambda \mathbb{E}_{ p({\bar{x}})}  [(||\nabla_{\bar{x}}D_{\theta}(\bar{x})||_2 - 1)^2],
\end{split}
\end{equation}
where $\lambda$ is the gradient penalty coefficient and $p(\bar{x})$ is the distribution sampled uniformly along the straight line between $p_{\text{data}}(x)$ and $p_G$. Following~\citet{karras2017progressive,gulrajani2017improved}, we set $\lambda=10.0$ in our experiments.

$(3)$ Hinge loss:\\
The hinge loss-based GAN loss function is given by
\begin{align}
 \label{eq:hinge}
    \underset{\theta}{\mbox{max}} \;& \mathbb{E}_{p_{\text{data}}(x)} [\min (0, -1+D_{\theta}(x))] \notag \\
    &+\mathbb{E}_{p(z)} [\min(0, -1-D_{\theta}(G_{\phi}(z)))];\\
    \underset{\phi}{\mbox{min}}\; \;&  -\mathbb{E}_{p(z)} [D_{\theta}(G_{\phi}(z))].
\end{align}
We set the learning rate to be 0.003 for results in Table~\ref{tab:loss-cmp}.

\section{More Generated Images}
\begin{figure*}[!ht]
    \centering
    \includegraphics[width=0.9\textwidth]{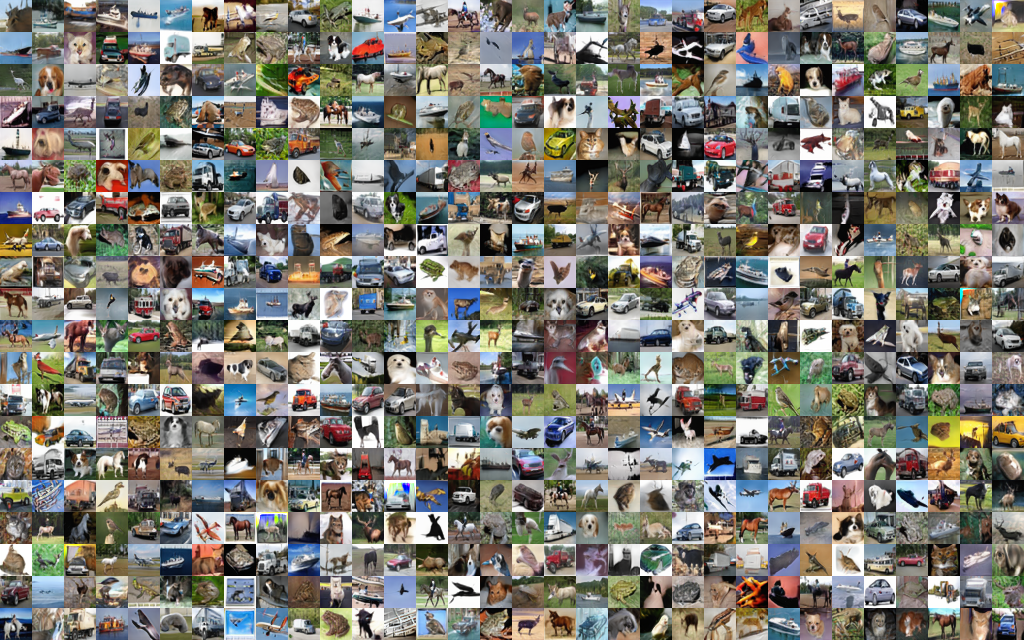}

    \caption{Generated images by the model trained on CIFAR-10 dataset without labels.}
    \label{fig:cifar10u}
\end{figure*}

\begin{figure*}[!ht]
    \centering
    \includegraphics[width=0.9\textwidth]{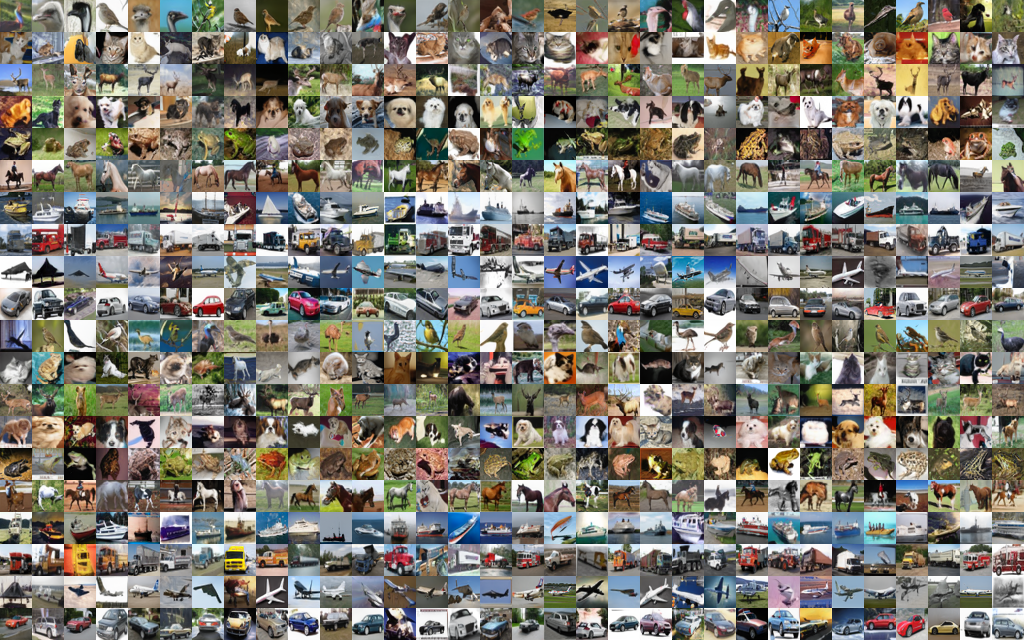}
    \caption{Generated images by the model trained on CIFAR-10 dataset with labels.}
    \label{fig:cifar10c}
\end{figure*}

\begin{figure*}[!ht]
    \centering
    \includegraphics[width=0.9\textwidth]{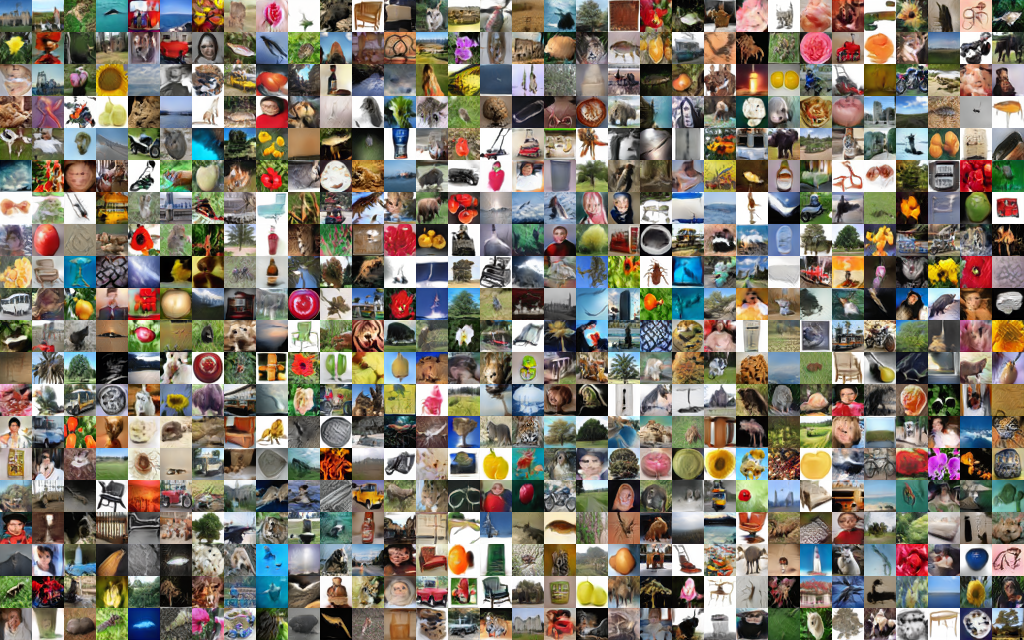}
    \caption{Generated images by the model trained on CIFAR-100 dataset without labels.}
    \label{fig:cifar100}
\end{figure*}

\begin{figure*}[!ht]
    \centering
    \includegraphics[width=0.9\textwidth]{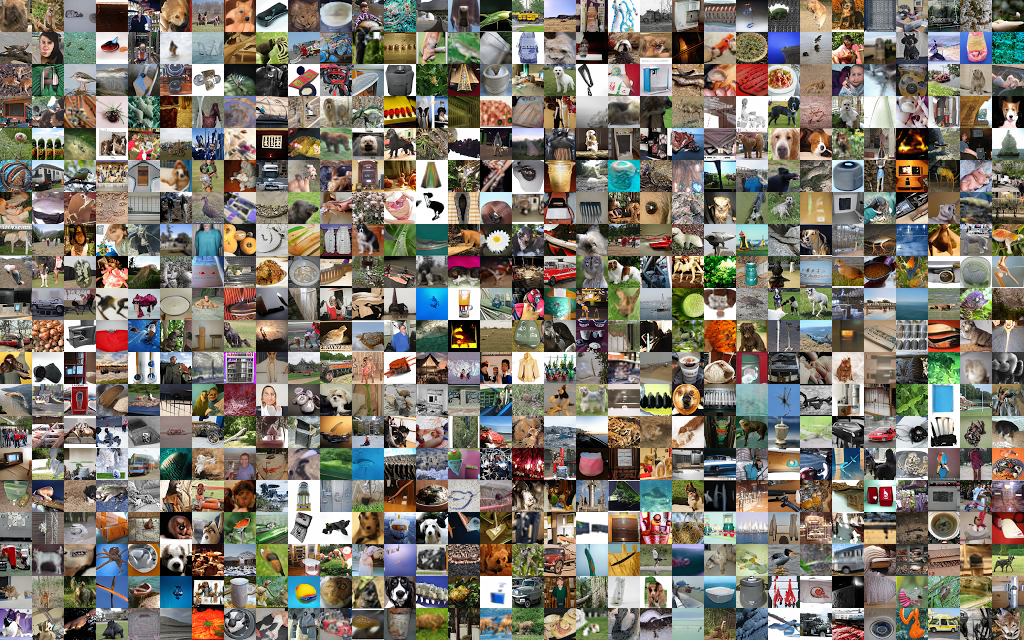}
    \caption{Generated images by the model trained on ImageNet ($32\times32$) without labels.}
    \label{fig:imagenet32}
\end{figure*}

\begin{figure*}[!ht]
    \centering
    \includegraphics[width=0.9\textwidth]{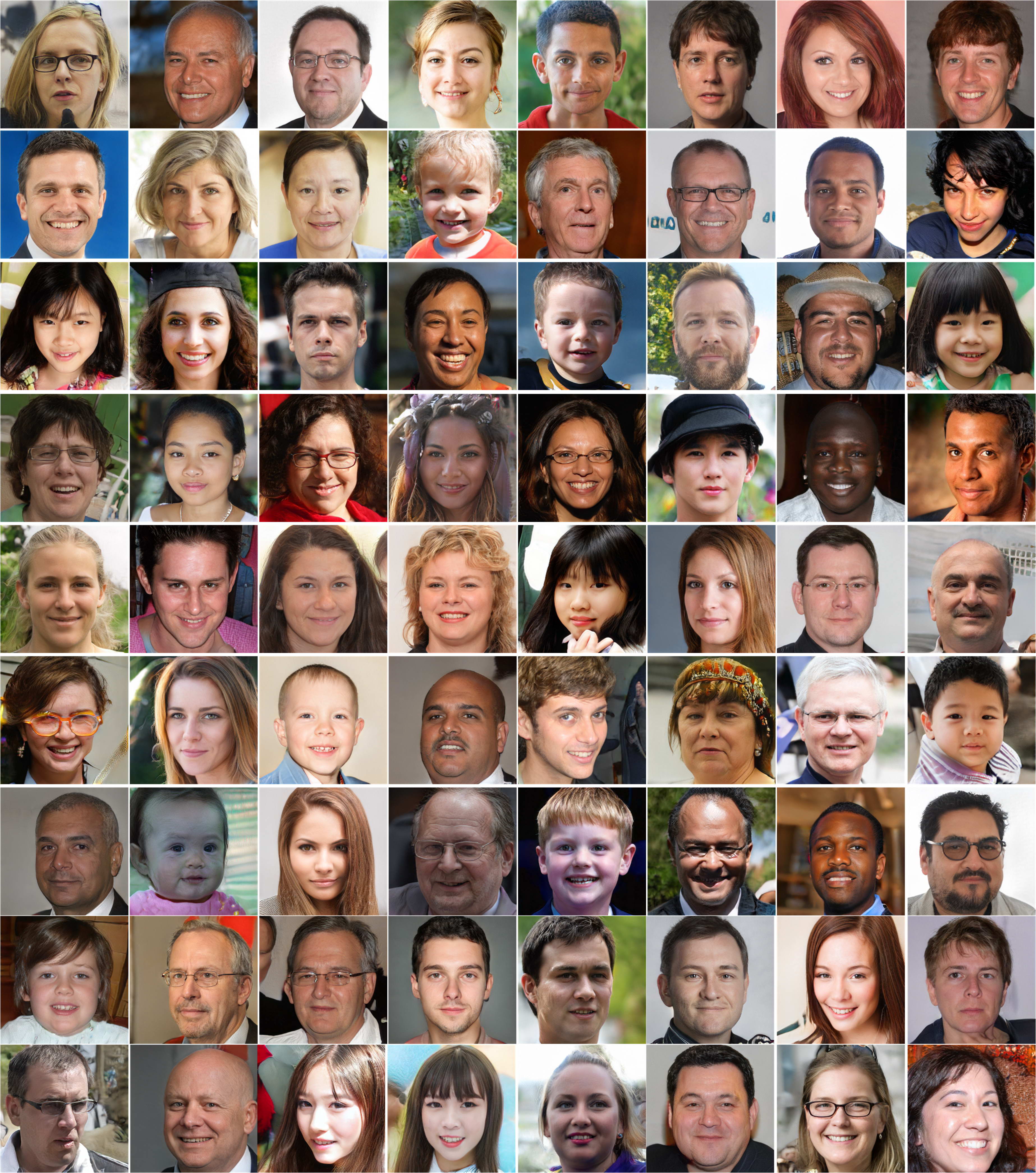}
    \caption{Generated images by the model trained on FFHQ dataset ($256\times256$).}
    \label{fig:ffhq256}
\end{figure*}

\begin{figure*}[!ht]
    \centering
    \includegraphics[width=0.99\textwidth]{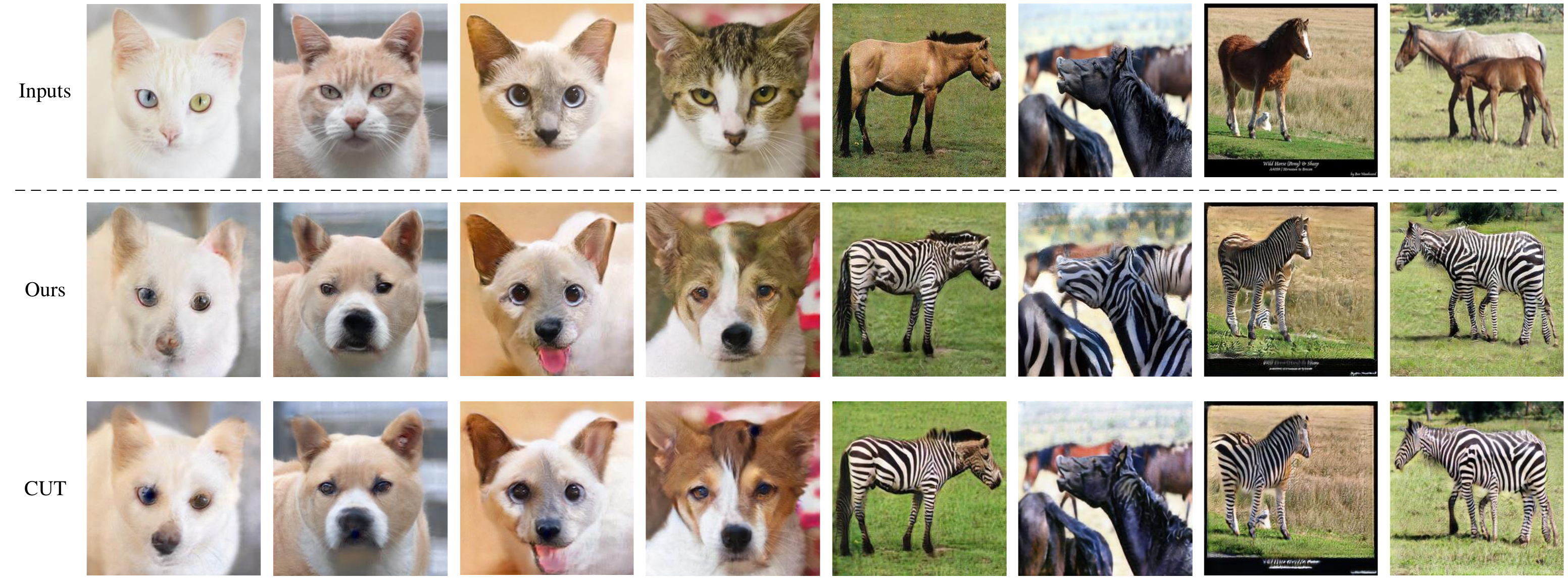}
    \caption{More results of one-sided unpaired image-to-image translation ($256\times256$).}
    \label{fig:i2i-supp}
\end{figure*}

In Figures~\ref{fig:cifar10u}~to~\ref{fig:ffhq256}, we present more qualitative results for image generation by our models trained on datasets CIFAR-10 (with and without labels), CIFAR-100, ImageNet $(32\times32)$, and FFHQ $(256\times256)$, respectively. Figure \ref{fig:i2i-supp} displays more qualitative results by our model and a baseline for the task of one-sided unpaired image-to-image translation.

\end{document}